\pdfoutput=1

\documentclass[11pt]{article}

\usepackage{emnlp2021}

\usepackage{times}
\usepackage{latexsym}
\usepackage{graphicx}
\usepackage{subfigure}

\usepackage[T1]{fontenc}

\usepackage[utf8]{inputenc}

\usepackage{microtype}

\usepackage{array}
\usepackage{pifont}
\usepackage{tabularx}
\usepackage{adjustbox}
\usepackage{multirow}
\usepackage{enumitem}
\usepackage{xspace}
\usepackage{tcolorbox}
\usepackage{booktabs,amsfonts,dcolumn}
\usepackage{hyperref}
\usepackage{url}
\usepackage{amsmath,amsthm,amsfonts,amssymb,bm,stmaryrd}
\usepackage[noorphans,vskip=0.75ex,leftmargin=2ex]{quoting}
\title{Simple Entity-Centric Questions Challenge Dense Retrievers}

\author{
  Christopher Sciavolino$^*$ \quad Zexuan Zhong$^*$ \quad Jinhyuk Lee \quad Danqi Chen\\
  Department of Computer Science, Princeton University\\
  \texttt{\{cds3,zzhong,jinhyuklee,danqic\}@cs.princeton.edu}
}

\newcommand\ti[1]{\textit{#1}}

\newcommand\tf[1]{\textbf{#1}}

\newcommand{\entity}{\text{[E]}}

\newcommand{\tableindent}{~~}

\newcommand{\eqa}{EntityQuestions}
\newcommand{\bornrel}{{place-of-birth}}
\newcommand{\creatorrel}{{creator}}
\newcommand{\headquarterrel}{{headquarter}}

\begin{document}
\maketitle

\renewcommand{\thefootnote}{\fnsymbol{footnote}}
\footnotetext[1]{The first two authors contributed equally.}
\renewcommand{\thefootnote}{\arabic{footnote}}


\begin{abstract}
Open-domain question answering has exploded in popularity recently due to the success of dense retrieval models, which have surpassed sparse models using only a few supervised training examples. However, in this paper, we demonstrate current dense models are not yet the holy grail of retrieval.
We first construct {\eqa}, a set of simple, entity-rich questions based on facts from Wikidata (e.g., ``Where was Arve Furset born?''), and observe that dense retrievers drastically underperform sparse methods.
We investigate this issue and uncover that dense retrievers can only generalize to common entities unless the question pattern is explicitly observed during training.
We discuss two simple solutions towards addressing this critical problem.
First, we demonstrate that data augmentation is unable to fix the generalization problem.
Second, we argue a more robust passage encoder helps facilitate better question adaptation using specialized question encoders.
We hope our work can shed light on the challenges in creating a robust, universal dense retriever that works well across different input distributions.\footnote{Our dataset and code are publicly available at \url{https://github.com/princeton-nlp/EntityQuestions}.}
\end{abstract}


\section{Introduction}
Recent dense passage retrievers outperform traditional sparse retrieval methods like TF-IDF and  BM25~\cite{robertson2009probabilistic} by a large margin on popular question answering datasets~(\citealt{lee2019latent}, \citealt{guu2020realm}, \citealt{karpukhin2020dense}, \citealt{xiong2021approximate}).
These dense models are trained using supervised datasets and the dense passage retriever (DPR) model~\cite{karpukhin2020dense} demonstrates that only training 1,000 supervised examples on top of BERT~\cite{devlin2019bert} already outperforms BM25, making it very appealing in practical use. In this work, we argue that dense retrieval models are not yet robust enough to replace sparse methods, and investigate some of the key shortcomings dense retrievers still face.

We first construct {\eqa}, an evaluation benchmark of simple, entity-centric questions like ``Where was Arve Furset born?'', and show dense retrieval methods generalize very poorly.
As shown in Table~\ref{tab:intro}, a DPR model trained on either a single dataset Natural Questions (NQ)~\cite{kwiatkowski2019natural} or a combination of common QA datasets drastically underperforms the sparse BM25 baseline (49.7\% vs 72.0\% on average), with the gap on some question patterns reaching $60\%$ absolute!


\begin{table}[!t]
    \centering
    \setlength{\tabcolsep}{0.1cm}
    \resizebox{1.0\columnwidth}{!}{
    \begin{tabular}{p{6cm}ccc} 
    \toprule
        & \tf{DPR} & \tf{DPR} & \tf{BM25} \\
        & (NQ) & (multi) & - \\
    \midrule 
        {Natural Questions} & {\tf{80.1}} & {79.4} & {64.4} \\
     \midrule
    {{\eqa}} (this work) & 49.7 & 56.7 & \tf{72.0} \\
    \tableindent\tableindent What is the capital of {\entity}? & 77.3 & 78.9 & \tf{90.6} \\
    \tableindent\tableindent  Who is {\entity} married to? & 35.6 & 48.1 & \tf{89.7} \\
    \tableindent\tableindent  Where is the headquarter of {\entity}? & 70.0 & 72.0 & \tf{85.0} \\
    \tableindent\tableindent  Where was {\entity} born? & 25.4 & 41.8 & \tf{75.3} \\
    \tableindent\tableindent  Where was {\entity} educated? & 26.4 & 41.8 & \tf{73.1} \\
    \tableindent\tableindent  Who was {\entity} created by? & 54.1 & 57.7 & \tf{72.6} \\
    \tableindent\tableindent  Who is {\entity}'s child? & 19.2 & 33.8 & \tf{85.0} \\
    \tableindent\tableindent  (17 more types of questions) & \ldots & \ldots & \ldots \\

    \bottomrule
    \end{tabular}
    }
    \caption{Top-20 retrieval accuracy for dense and sparse retrieval models on Natural Questions~\cite{kwiatkowski2019natural} and our {\eqa} along with a set of sampled questions (full results in Appendix~\ref{app:full_results}). We test two DPR models: (1) trained on NQ only; (2) trained on 4 datasets (NQ, TQA, WebQ, TREC) combined.\protect\footnotemark}
    \label{tab:intro}
    \vspace{-1em}
\end{table}

\footnotetext{
    NQ: Natural Questions~\cite{kwiatkowski2019natural}, TQA: TriviaQA~\cite{joshi2017triviaqa}, WebQ: WebQuestions~\cite{berant2013semantic}, CuratedTREC~\cite{baudivs2015modeling}.  We use the pre-trained models from \url{https://github.com/facebookresearch/DPR}.
    }

Based on these results, we perform a deep dive into why a single dense model performs so poorly on these simple questions.
We decouple the two distinct aspects of these questions: the entities and the question pattern, and identify what about these questions gives dense models such a hard time.
We discover the dense model is only able to successfully answer questions based on common entities, quickly degrading on rarer entities.
We also observe that dense models can generalize to unseen entities only when the question pattern is explicitly observed during training.

We end with two investigations of practical solutions towards addressing this crucial problem.
First, we consider data augmentation and analyze the trade-off between single- and multi-task fine-tuning.
Second, we consider a fixed passage index and fine-tune specialized question encoders, leading to memory-efficient transfer to new questions.

We find that data augmentation, while able to close gaps on a single domain, is unable to consistently improve performance on unseen domains.
We also find that building a robust passage encoder is crucial in order to successfully adapt to new domains.
We view this study as one important step towards building universal dense retrieval models.


\section{Background and Related Work}
\paragraph{Sparse retrieval}
Before the emergence of dense retrievers, traditional sparse retrievers such as TF-IDF or BM25 were the de facto method in open-domain question-answering systems~\cite{chen2017reading,yang2019end}.
These sparse models measure similarity using weighted term-matching between questions and passages and do not train on a particular data distribution.
It is well-known that sparse models are great at lexical matching, but fail to capture synonyms and paraphrases.

\paragraph{Dense retrieval} On the contrary, dense models~\cite{lee2019latent,karpukhin2020dense,guu2020realm} measure similarity using learned representations from supervised QA datasets, leveraging pre-trained language models like BERT.
In this paper, we use the popular dense passage retriever (DPR) model~\cite{karpukhin2020dense} as our main evaluation,\footnote{The detailed experimental settings are in Appendix~\ref{app:exp-details}.} and we also report the evaluation of REALM~\cite{guu2020realm} in Appendix~\ref{app:full_results}. DPR models the retrieval problem using two encoders, namely the question and the passage encoders, initialized using BERT. DPR uses a contrastive objective during training, with in-batch negatives and hard negatives mined from BM25. During inference, a pre-defined large set of passages (e.g., 21-million passages in English Wikipedia) are encoded and pre-indexed---for any test question, the top passages with the highest similarity scores are returned. Recently, other advances have been made in improving dense retrieval, including incorporating better hard negatives~\cite{xiong2021approximate,qu2021rocketqa}, or fine-grained phrase retrieval~\cite{lee2021learning}. We leave them for future investigation.

\paragraph{Generalization problem}
Despite the impressive in-domain performance of dense retrievers, their capability of generalizing to unseen questions still remains relatively under-explored.
Recently, \newcite{lewis2020question} discover that there is a large overlap between training and testing sets on popular QA benchmarks, concluding that current models tend to memorize training questions and perform significantly worse on non-overlapping questions.
AmbER~\cite{chen2021evaluating} test sets are designed to study the entity disambiguation capacities of passage retrievers and entity linkers. They find models perform much worse on rare entities compared to common entities.
Similar to this work, our results show dense retrieval models generalize poorly, especially on rare entities. We further conduct a series of analyses to dissect the problem and investigate potential approaches for learning robust dense retrieval models.
Finally, another concurrent work~\cite{thakur2021beir} introduces the BEIR benchmark for zero-shot evaluation of retrieval models and shows that dense retrieval models underperform BM25 on most of their datasets.


\begin{figure*}[!t]
\centering
\hspace{-2em}
\begin{minipage}[b]{.41\textwidth}
    \centering
    \includegraphics[width=1.1\columnwidth]{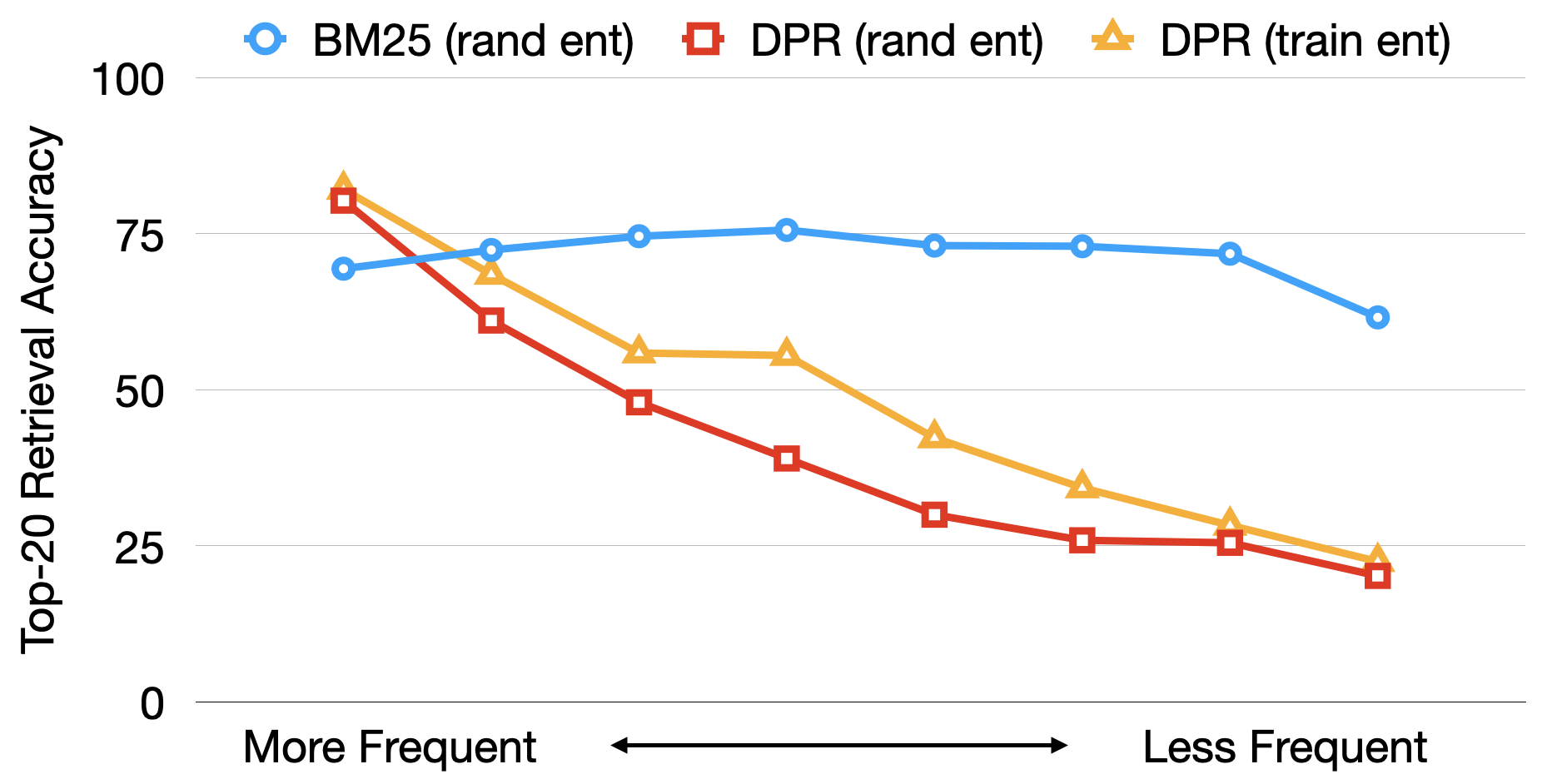}
    \vspace{-2em}
    \caption*{~~~~~~~~(a) {\bornrel}}
\end{minipage}\qquad
\begin{minipage}[b]{.41\textwidth}
    \centering
    \includegraphics[width=1.1\columnwidth]{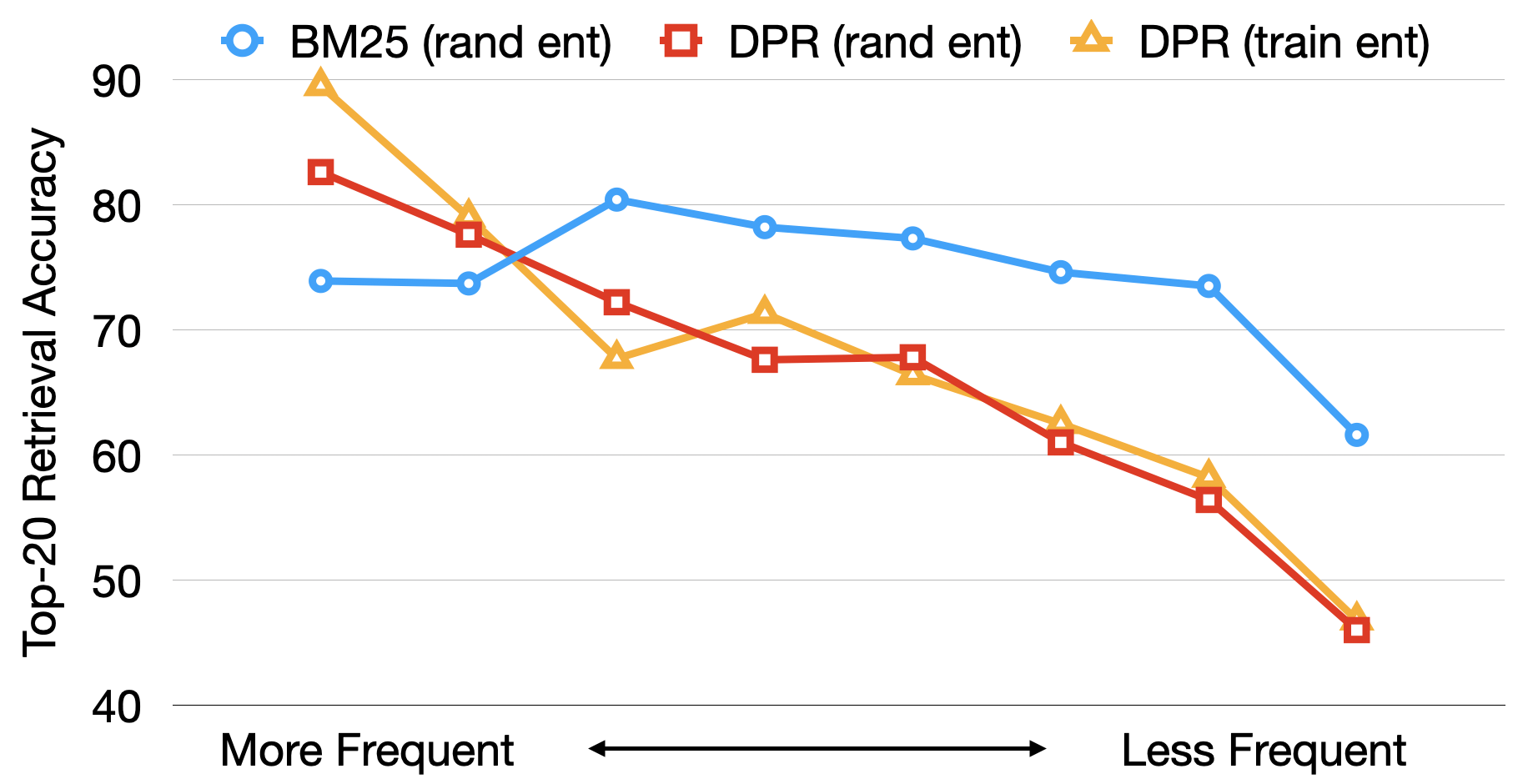}
    \vspace{-2em}
    \caption*{~~~~~~~~(b) {\creatorrel}}
\end{minipage}\qquad

\vspace{-0.5em}
\caption{
Top-20 retrieval accuracy on (a) \ti{\bornrel} questions, (b) \ti{\creatorrel} questions, grouped by the entity's frequency in Wikipedia.
    We group entities into 8 buckets according to their frequency in Wikipedia.
    \textit{rand ent}: randomly selected entities from Wikidata; \textit{train ent}: entities in the NQ training set.
}
\label{fig:freq-analysis}
\end{figure*}

\section{\eqa}
In this section, we build a new benchmark {\eqa}, a set of simple, entity-centric questions and compare dense and sparse retrievers.
\paragraph{Dataset collection}
We select 24 common relations from Wikidata~\cite{wikidata2014vrandecic} and convert fact (\textit{subject}, \textit{relation}, \textit{object}) triples into natural language questions using manually defined templates (Appendix~\ref{app:full_results}).
To ensure the converted natural language questions are answerable from Wikipedia, we sample triples from the T-REx dataset~\cite{trex2018elsahar}, where triples are aligned with a sentence as evidence in Wikipedia.
We select relations following the criteria:
(1) there are enough triples ($>$2k) in the T-REx;
(2) it is easy enough to formulate clear questions for the relation;
(3) we do not select relations with only a few answer candidates (e.g., \textit{gender}), which may cause too many false negatives when we evaluate the retriever;
(4) we include both person-related relations (e.g., \textit{place-of-birth}) and non-person relations (e.g., \textit{headquarter}).
For each relation, we randomly sample up to 1,000 facts to form the evaluation set.
We report the macro-averaged accuracy over all relations of {\eqa}.

\paragraph{Results}
We evaluate DPR and BM25 on the {\eqa} dataset and report results in Table~\ref{tab:intro} (see full results and examples in Appendix~\ref{app:full_results}).
DPR trained on NQ significantly underperforms BM25 on almost all sets of questions.
For example, on the question ``Where was {\entity} born?'', BM25 outperforms DPR by $49.9\%$ absolute using top-20 retrieval accuracy.\footnote{For our entire analysis, we consider top-20 retrieval accuracy for brevity. However, trends still hold for top-1, top-5, and top-100 retrieval accuracy.} Although training DPR on multiple datasets can improve the performance (i.e., from $49.7\%$ to $56.7\%$ on average), it still clearly pales in comparison to BM25. We note the gaps are especially large on questions about person entities.

In order to test the generality of our findings, we also evaluate the retrieval performance of REALM~\cite{guu2020realm} on {\eqa}. Compared to DPR, REALM adopts a pre-training task called salient span masking (SSM), along with an inverse cloze task from \newcite{lee2019latent}.
We include the evaluation results in Appendix~\ref{app:full_results}.\footnote{We cannot directly compare the retrieval accuracy of REALM to DPR, as the REALM index uses 288 BPE token blocks while DPR uses 100 word passages.}
We find that REALM still scores much lower than BM25 over all relations  ($19.6\%$ on average).
This suggests that incorporating pre-training tasks such as SSM still does not solve the generalization problem on these simple entity-centric questions.


\section{Dissecting the Problem: Entities vs. Question Patterns}
In this section, we investigate why dense retrievers do not perform well on these questions. Specifically, we want to understand whether the poor generalization should be attributed to (a) novel entities, or (b) unseen question patterns. To do this, we study DPR trained on the NQ dataset and evaluate on three representative question templates: \ti{\bornrel}, \ti{\headquarterrel}, and \ti{\creatorrel}.\footnote{The question templates for these relations are: {\bornrel}: ``Where was {\entity} born?''; {\headquarterrel}: ``Where is the headquarters of {\entity}?''; {\creatorrel}: ``Who was {\entity} created by?''.}

\subsection{Dense retrievers exhibit popularity bias}
We first determine how the entity {\entity} in the question affects DPR's ability to retrieve relevant passages.
To do this, we consider all triples in Wikidata that are associated with a particular relation, and order them based on frequency of the subject entity in Wikipedia. In our analysis, we use the Wikipedia hyperlink count as a proxy for an entity's frequency.
Next, we group the triples into $8$ buckets such that each bucket has approximately the same cumulative frequency.

Using these buckets, we consider two new evaluation sets for each relation. The first (denoted ``rand ent'') randomly samples at most 1,000 triples from each bucket. The second (denoted ``train ent'') selects all triples within each bucket that have subject entities observed in questions within the NQ training set, as identified by ELQ~\cite{li-etal-2020-efficient}.

We evaluate DPR and BM25 on these evaluation sets and plot the top-20 accuracy in Figure~\ref{fig:freq-analysis}.
DPR performs well on the most common entities but quickly degrades on rarer entities, while BM25 is less sensitive to entity frequency.
It is also notable that DPR performs generally better on entities seen during NQ training than on randomly selected entities. This suggests that DPR representations are much better at representing the most common entities as well as entities observed during training.


\begin{table}[!t]
    \centering
    \resizebox{1\columnwidth}{!}{
    \begin{tabular}{l  c  c  c}
        \toprule
         & \tf{p-of-birth} & \tf{\headquarterrel} & \tf{\creatorrel} \\
        \midrule
        DPR-NQ & 25.4 & 70.0 & 54.1 \\
        \tableindent FT & 73.9 & \tf{84.0} & \tf{80.0} \\
        \tableindent FT w/ similar & \tf{74.7} & 79.9 & 76.2 \\
        \midrule
        \tableindent FT OnlyP & \tf{72.8} & \tf{84.2} & \tf{78.0} \\
        \tableindent FT OnlyQ & 45.4 & 72.8 & 73.4 \\
        \midrule
        BM25 & 75.3 & 85.0 & 72.6 \\
        \bottomrule
    \end{tabular}
    }
    \caption{Top-20 retrieval accuracy on {\eqa} test sets after fine-tuning. FT: fine-tuning on each individual question pattern. \textit{w/ similar}: fine-tuning on a similar, semantically equivalent question pattern.
    \ti{OnlyP} and \ti{OnlyQ}: fixing the weights of the question encoder and only updating the passage encoder, or vice versa.
    }
    \vspace{-1.2em}
    \label{tab:p19_finetune_analysis}
\end{table}


\subsection{Observing questions helps generalization}
\label{sec:entity_generalization}
We next investigate whether DPR generalizes to unseen entities when trained on the question pattern.
For each relation considered, we build a training set with at most $8,000$ triples.
We ensure no \ti{tokens} from training triples overlap with tokens from triples in the corresponding test set.
In addition to using the question template used during evaluation to generate training questions, we also build a training set based on a syntactically different but semantically equal question template.\footnote{
{\bornrel}: ``What is the birthplace of {\entity}?''; {\headquarterrel}: ``Where is {\entity} headquartered?''; {\creatorrel}: ``Who is the creator of {\entity}?''.
}
We fine-tune DPR models on the training set for each relation and test on the evaluation set of {\eqa} for the particular relation and report results in Table~\ref{tab:p19_finetune_analysis}.

\begin{figure*}[t]
\centering
\hspace{-3em}
\begin{minipage}[b]{.43\textwidth}
    \centering
    \includegraphics[width=1.1\columnwidth]{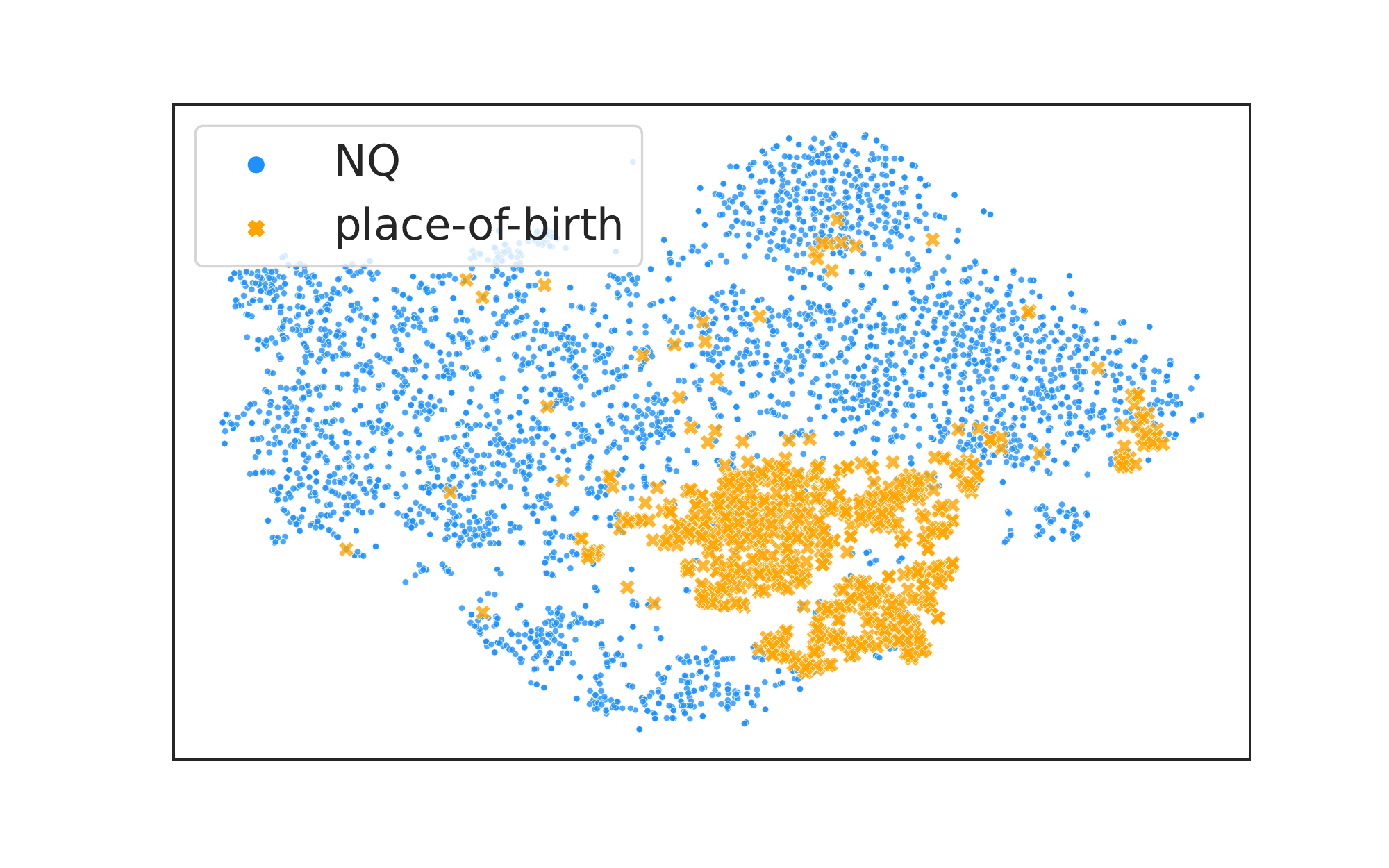}
    \vspace{-3em}
    \caption*{~~~~~~~(a)}
\end{minipage}\qquad
\begin{minipage}[b]{.43\textwidth}
    \centering
    \includegraphics[width=1.1\columnwidth]{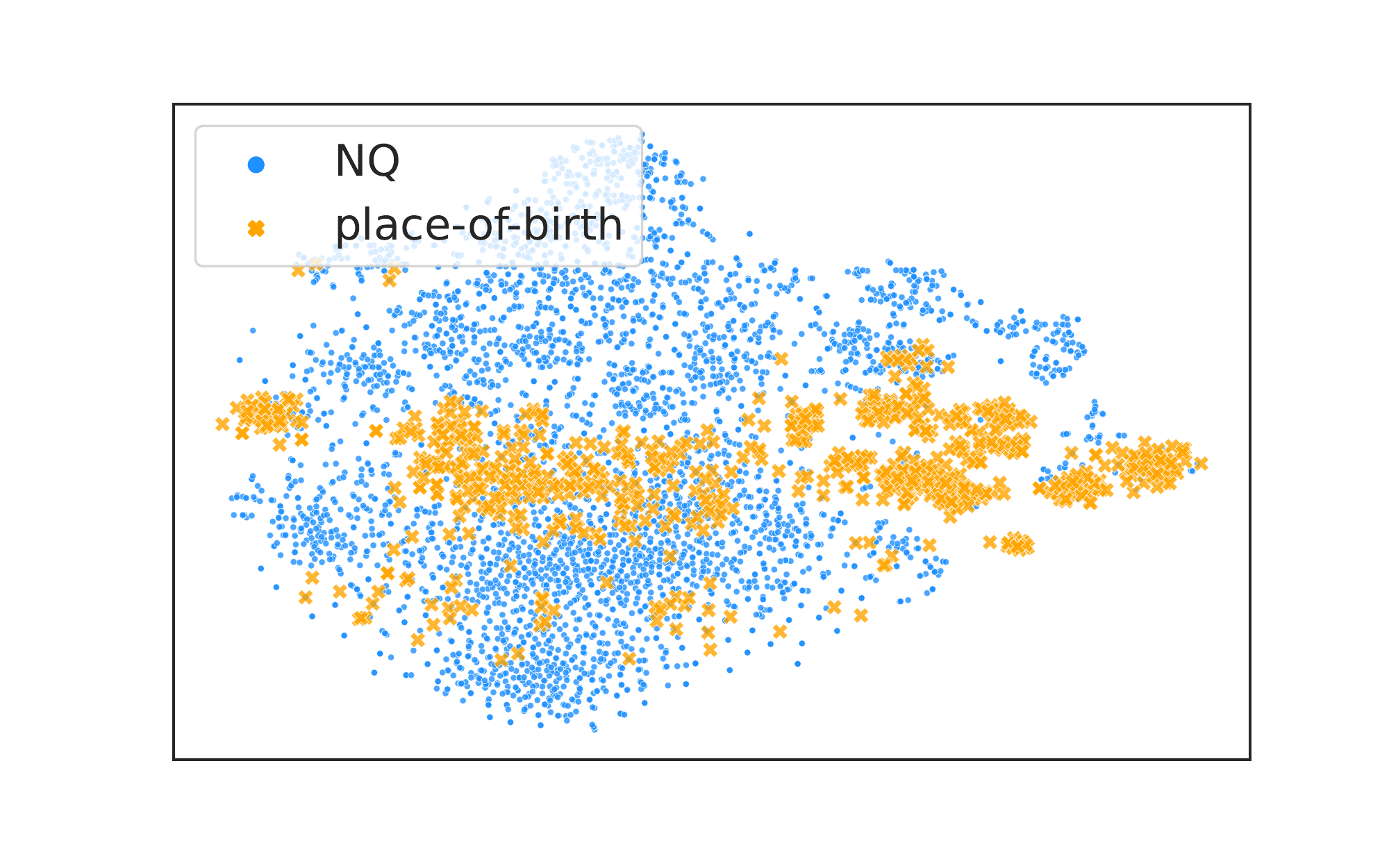}
    \vspace{-3em}
    \caption*{~~~~~~~(b)}
\end{minipage}
\caption{
Visualization of positive passage embeddings returned by DPR before and after fine-tuning on the \ti{place-of-birth} questions.
(a): Positive passage embeddings returned by DPR trained on NQ; (b) Positive passage embeddings returned by DPR after fine-tuning.
}
\label{fig:tsne_plots}
\end{figure*}

Clearly, observing the question pattern during training allows DPR to generalize well on unseen entities.
On all three relations, DPR can match or even outperform BM25 in terms of retrieval accuracy.
Training on the equivalent question pattern achieves comparable performance to the exact pattern, showing dense models do not rely on specific phrasing of the question.
We also attempt fine-tuning the question encoder and passage encoder separately.
As shown in Table~\ref{tab:p19_finetune_analysis}, surprisingly, there is a significant discrepancy between only training the passage encoder (OnlyP) and only training the question encoder (OnlyQ): for example, on {\bornrel}, DPR achieves $72.8\%$ accuracy with the fine-tuned passage encoder, while it achieves $45.4\%$ if only the question encoder is fine-tuned.
This suggests that passage representations might be the culprit for model generalization.

To understand what passage representations have learned from fine-tuning, we visualize the DPR passage space before and after fine-tuning using t-SNE~\cite{van2008visualizing}.
We plot the representations of positive passages sampled from NQ and \ti{\bornrel} in Figure~\ref{fig:tsne_plots}.
Before fine-tuning, positive passages for \ti{\bornrel} questions are clustered together.
Discriminating passages in this clustered space is more difficult using an inner product, which explains why only fine-tuning the question encoder yields minimal gains.
After fine-tuning, the passages are distributed more sparsely, making differentiation much easier.


\section{Towards Robust Dense Retrieval}
\label{sec:robust_dense}
Equipped with a clear understanding of the issues, we explore some simple techniques aimed at fixing the generalization problem. 

\paragraph{Data augmentation}
We first explore whether fine-tuning on questions from a \ti{single} {\eqa} relation can help generalize on the full set of {\eqa} as well as other QA datasets such as NQ. We construct a training set of questions for a single relation and consider two training regimes: one where we fine-tune on relation questions alone; and a second where we fine-tune on both relation questions and NQ in a multi-task fashion. We perform this analysis for three relations and report top-20 retrieval accuracy in Table~\ref{tab:data_augmentation_entityqa}.


\begin{table}[!t]
    \centering
    \resizebox{1.0\columnwidth}{!}{
    \begin{tabular}{l ccc}
        \toprule
        & \tf{NQ} & \tf{Rel} & \tf{EntityQ.} \\
        \midrule
        DPR-NQ & \tf{80.1} & 25.4 & 49.7 \\
        + FT \textit{p-of-birth} & 62.8 & \tf{74.3} & \tf{56.2} \\
        + FT NQ $\cup$ \textit{p-of-birth} & 70.8 & 52.0 & 47.4 \\
        \midrule
        DPR-NQ & \tf{80.1} & 70.0 & 49.7 \\
        + FT \textit{headquarter} & 71.6 & 80.3 & \tf{53.3}\\
        + FT NQ $\cup$ \textit{headquarter} & 75.1 & \tf{81.3} & 49.5 \\
        \midrule
        DPR-NQ & \tf{80.1} & 54.1 & 49.7 \\
        + FT \textit{creator} & 70.8 & \tf{80.8} & \tf{52.3} \\
        + FT NQ $\cup$ \textit{creator} & 72.6 & 72.3 & 44.1 \\
        \midrule
        BM25 & 64.4 & - & \tf{72.0} \\
        \bottomrule
    \end{tabular}
    }
    \caption{Top-20 retrieval accuracy on NQ and {\eqa}.
    \textit{FT}: fine-tuning.
    \textit{Rel}: the performance on the relation that is used during fine-tuning.
    }
    \vspace{-1.0em}
    \label{tab:data_augmentation_entityqa}
\end{table}


We find that fine-tuning only on a single relation improves {\eqa} meaningfully, but degrades performance on NQ and still largely falls behind BM25 on average. When fine-tuning on both relation questions and NQ together, most of the performance on NQ is retained, but the gains on {\eqa} are much more muted.
Clearly, fine-tuning on one type of entity-centric question does not necessarily fix the generalization problem for other relations.
This trade-off between accuracy on the original distribution and improvement on the new questions presents an interesting tension for universal dense encoders to grapple with.

\paragraph{Specialized question encoders}
While it is challenging to have one retrieval model for all unseen question distributions, we consider an alternative approach of having a single passage index and adapting specialized question encoders.
Since the passage index is fixed across different question patterns and cannot be adapted using fine-tuning, having a robust passage encoder is crucial.

We compare two DPR passage encoders: one based on NQ and the other on the PAQ dataset~\cite{lewis2021paq}.\footnote{PAQ dataset sampling scheme is described in Appendix~\ref{app:exp-details}.}
We expect the question encoder trained on PAQ is more robust
because (a) 10M passages are sampled in PAQ, which is arguably more varied than NQ, and (b) all the plausible answer spans are identified using automatic tools.
We fine-tune a question encoder for each relation in {\eqa}, keeping the passage encoder fixed.
As shown in Table~\ref{tab:passage_encoders},\footnote{Per-relation accuracy can be found in Appendix~\ref{app:per-relation}.} fine-tuning the encoder trained on PAQ improves performance over fine-tuning the encoder trained on NQ.
This suggests the DPR-PAQ encoder is more robust,
{ }nearly closing the gap with BM25 using a single passage index. 
We believe constructing a robust passage index is an encouraging avenue for future work towards a more general retriever.


\begin{table}[!t]
    \centering
    \resizebox{0.85\columnwidth}{!}{
    \begin{tabular}{l c}
        \toprule
        & \tf{EntityQ.}\\
        \midrule
        DPR-NQ$^\dagger$  & 45.1 \\
        + Per-relation FT (OnlyQ) & \tf{61.6} \\
        + {\eqa} FT (OnlyQ) & 53.0 \\
        \midrule
        DPR-PAQ$^\dagger$ & 59.3 \\
        + Per-relation FT (OnlyQ) & \tf{68.4} \\
        + {\eqa} FT (OnlyQ) & 65.4 \\
        \midrule
        BM25 & \tf{72.0} \\
        \bottomrule
    \end{tabular}
    }
    \caption{Averaged top-20 retrieval accuracy on \eqa. We fix the passage encoder and fine-tune the question encoder. \textit{Per-relation FT}: fine-tuning an individual question encoder for each relation, \textit{{\eqa} FT}: fine-tuning on all questions in {\eqa}. $^\dagger$: we re-train the models from scratch.
    }
    \vspace{-0.8em}
    \label{tab:passage_encoders}
\end{table}



\section{Conclusion}
In this study, we show that DPR significantly underperforms BM25 on {\eqa}, a dataset of simple questions based on facts mined from Wikidata.
We derive key insights about why DPR performs so poorly on this dataset.
We learn that DPR remembers robust representations for common entities, but struggles to differentiate rarer entities without training on the question pattern.

We suggest future work in incorporating entity memory into dense retrievers to help differentiate rare entities.
Several recent works demonstrate retrievers can easily learn dense representations for a large number of Wikipedia entities~\cite{wu-etal-2020-scalable,li-etal-2020-efficient}, or directly generate entity names in an autoregressive manner~\cite{de2021autoregressive}.
DPR could also leverage entity-aware embedding models like EaE~\cite{fevry-etal-2020-entities} or LUKE~\cite{yamada2020luke} to better recall long-tail entities.

\newpage

\section*{Acknowledgements}
We thank the members of the Princeton NLP group for helpful discussion and valuable feedback.
This research is supported by gift awards from Apple and Amazon.


\section*{Ethical Considerations}
Our proposed dataset, {\eqa}, is constructed by sampling (subject, relation, object) triples from Wikidata, which is dedicated to the public domain under the Creative Commons CC0 License.
In general, machine learning has the ability to amplify biases presented implicitly and explicitly in the training data.
Models that we reference in our study are based on BERT, which has been shown to learn and exacerbate stereotypes during training~(e.g., \citealt{kurita-etal-2019-measuring}, \citealt{tan-2019-assessing}, \citealt{nadeem2020stereoset}).
We further train these models on Wikidata triples, which again has the potential to amplify harmful and toxic biases.

In the space of open-domain question answering, deployed systems leveraging biased pre-trained models like BERT will likely be less accurate or biased when asked questions related to stereotyped and marginalized groups. We acknowledge this fact and caution those who build on our work to consider and study this implication before deploying systems in the real world.

\bibliography{ref_v2}
\bibliographystyle{acl_natbib}

\clearpage
\newpage
\appendix

\begin{table}[!t]
    \centering
    \setlength{\tabcolsep}{0.1cm}
    \resizebox{1.0\columnwidth}{!}{
    \begin{tabular}{llccc} 
    \toprule
        \multicolumn{2}{l}{\tf{Relation}} & \tf{DPR} & \tf{DPR} & \tf{BM25} \\
         & & (NQ) & (multi) & \\
    \midrule
    P36 & What is the capital of {\entity}? & 77.3 & 78.9 & \tf{90.6} \\
    P407 & Which language was {\entity} written in? & 77.1 & 82.5 & \tf{86.2} \\
    P26 & Who is {\entity} married to? & 35.6 & 48.1 & \tf{89.7} \\
    P159 & Where is the headquarter of {\entity}? & 70.0 & 72.0 & \tf{85.0} \\
    P276 & Where is {\entity} located? & 74.9 & 77.3 & \tf{84.9} \\
    P40 & Who is {\entity}'s child? & 19.2 & 33.8 & \tf{85.0} \\
    P176 & Which company is {\entity} produced by? & 61.7 & 73.7 & \tf{81.0} \\
    P20 & Where did {\entity} die? & 34.4 & 45.1 & \tf{80.4} \\
    P112 & Who founded {\entity}? & 77.1 & 75.7 & \tf{81.2} \\
    P127 & Who owns {\entity}? & 60.7 & 63.8 & \tf{78.4} \\
    P19 & Where was {\entity} born? & 25.4 & 41.8 & \tf{75.3} \\
    P740 & Where was {\entity} founded? & 59.9 & 61.6 & \tf{74.4} \\
    P413 & What is {\entity} famous for? & \tf{75.7} & 71.5 & 74.3 \\
    P800 & What position does {\entity} play? & 19.0 & 33.9 & \tf{74.7} \\
    P69 & Where was {\entity} educated? & 26.4 & 41.8 & \tf{73.1} \\
    P50 & Who is the author of {\entity}? & 75.7 & \tf{77.8} & 73.0 \\
    P170 & Who was {\entity} created by? & 54.1 & 57.7 & \tf{72.6} \\
    P106 & What kind of work does {\entity} do? & 25.9 & 52.9 & \tf{71.2} \\
    P131 & Where is {\entity} located? & 45.7 & 44.2 & \tf{63.1} \\
    P17 & Which country is {\entity} located in? & 64.2 & \tf{67.7} & 61.5 \\
    P175 & Who performed {\entity}? & 47.6 & 51.5 & \tf{56.6} \\
    P136 & What type of music does {\entity} play? & 37.4 & 36.8 & \tf{48.7} \\
    P264 & What music label is {\entity} represented by? & 25.3 & 43.2 & \tf{45.6} \\
    P495 & Which country was {\entity} created in? & 21.6 & \tf{28.0} & 21.8 \\
    \midrule
    \multicolumn{2}{l}{\ti{Macro-Average}} & 49.7 & 56.7 & \tf{72.0} \\
    \multicolumn{2}{l}{\ti{Micro-Average}} & 49.5 & 56.6 & \tf{71.3} \\
    \bottomrule
    \end{tabular}
    }
    \caption{Top-20 retrieval accuracy (percentage of retrieved passages that contain the answer) for dense and sparse retrieval models on different sets of questions of {\eqa}.  We test two DPR models: (1) trained on NQ only; (2) trained on 4 datasets (NQ, TQA, WebQ, TREC) combined. }
    \label{tab:full_results}
\end{table}

\begin{table}[!t]
    \centering
    \setlength{\tabcolsep}{0.1cm}
    \resizebox{1.0\columnwidth}{!}{
    \begin{tabular}{llrr} 
    \toprule
      \multicolumn{2}{l}{\tf{Relation}} & \tf{REALM} & \tf{BM25} \\
    \midrule 
P36 & What is the capital of {\entity}? & 91.7 & \tf{91.9} \\
P407 & Which language was {\entity} written in? & 81.9 & \tf{92.0} \\
P26 & Who is {\entity} married to? & 47.1 & \tf{90.0} \\
P159 & Where is the headquarter of {\entity}? & 70.4 & \tf{90.7} \\
P276 & Where is {\entity} located? & 77.1 & \tf{89.5} \\
P40 & Who is {\entity}'s child? & 39.7 & \tf{87.1} \\
P176 & Which company is {\entity} produced by? & 69.2 & \tf{83.2} \\
P20 & Where did {\entity} die? & 61.9 & \tf{89.2} \\
P112 & Who founded {\entity}? & 77.3 & \tf{85.5} \\
P127 & Who owns {\entity}? & 73.6 & \tf{84.4} \\
P19 & Where was {\entity} born? & 52.9 & \tf{90.3} \\
P740 & Where was {\entity} founded? & 50.9 & \tf{77.5} \\
P413 & What position does {\entity} play? & 53.8 & \tf{90.4} \\
P800 & What is {\entity} famous for? & 45.3 & \tf{81.9} \\
P69 & Where was {\entity} educated? & 38.6 & \tf{84.1} \\
P50 & Who is the author of {\entity}? & \tf{77.2} & 76.2 \\
P170 & Who was {\entity} created by? & 56.8 & \tf{78.5} \\
P106 & What kind of work does {\entity} do? & 53.6 & \tf{83.4} \\
P131 & Where is {\entity} located? & 63.9 & \tf{86.8} \\
P17 & Which country is {\entity} located in? & 70.6 & \tf{76.0} \\
P175 & Who performed {\entity}? & 53.1 & \tf{65.8} \\
P136 & What type of music does {\entity} play? & 42.6 & \tf{53.4} \\
P264 & What music label is {\entity} represented by? & 53.2 & \tf{55.3} \\
P495 & Which country was {\entity} created in? & \tf{34.8} & 24.8 \\
\midrule
\multicolumn{2}{l}{\ti{Macro-Average}} & 59.9 & \tf{79.5} \\
\multicolumn{2}{l}{\ti{Micro-Average}} & 59.5 & \tf{79.0} \\
    \bottomrule
    \end{tabular}
    }
    \caption{Top-20 retrieval accuracy of REALM and BM25 on different sets of questions of {\eqa}. In this table, BM25 and REALM both only check for the answer in the passage text, excluding the title.}
    \label{tab:realm_results}
\end{table}


\begin{table*}[!t]
\small
\begin{tabular}{p{3cm}|p{11cm}}
\toprule
\tf{Question} & \tf{Retrieved results from DPR-NQ} \\
\midrule
Q: Where was Arve Furset born? & \tf{Gold}: [Title: Arve Furset] Arve Furset Arve Eilif Furset (born 5 December 1964 in \underline{Askvoll}, Western Norway) is a Norwegian composer, jazz musician (piano, keyboards) and music producer\\ 
& \tf{Top-1}: [Title: Gard Agdi] Gard Agdi ("Old Norse" Garðr Agði) appears in the legendary genealogies of "Hversu Noregr byggdist" as one of the three sons of \ldots\\
& \tf{Top-2}: [Title: Yrsa] kidnapped the queen for a while during which time he made her pregnant. Having returned to her kingdom, the queen bore a child, a girl which she named Yrsa after her dog. Yrsa was sent\ldots \\
& \tf{Top-3}: [Title: Arvid Noe] Sailor" and the anagram "Arvid Noe" to conceal his identity; his true name, Arne Vidar Røed, became known after his death. Røed began his career as a sailor in 1961, when he\ldots \\
\midrule
Q: Where was Mason Proper founded? & \tf{Gold}: [Title: Mason Proper] Mason Proper is an American rock band formed in \underline{Alpena}, Michigan, in 2004. The band consists of Jonathan Visger (singer), Matt\ldots\\
& \tf{Top-1}: [Title: George Mason] with the world. Thus, colonial Virginia initially developed few towns, since estates were largely self-sufficient, and could get what they needed without the need to purchase locally. Even the\ldots\\
& \tf{Top-2}: [Title: History of Freemasonry] so on through an elaborate path to Athelstan. This myth formed the basis for subsequent manuscript constitutions, all tracing masonry back to biblical times, and fixing its institutional\ldots\\
& \tf{Top-3}: [Title: Mason (surname)] Masullo, Masutti, Masutto, and several others. Mason (surname) Mason is an Italian, French Irish or English surname that refers to someone who did stonemasonry work, or it derives from\ldots \\
\midrule
Q: Who was Veep created by? & \tf{Gold}: [Title: Veep]  Veep is an American political satire comedy television series... created by Scottish writer \underline{Armando Iannucci} as an...\\
& \tf{Top-1}: [Title: Beep the Meep] (DWM \#250). The Meeps (although not Beep himself) appeared in the spin-off short story "Wish Upon A Star Beast" by Steve Lyons, published in the charity anthology "Perfect Timing". Beep\ldots \\
& \tf{Top-2}: [Title: Beep the Meep] Beep the Meep is a fictional alien who appeared in the "Doctor Who Weekly" comic strip based on the long-running British science fiction television series "Doctor Who". The cute and cuddly\ldots \\
& \tf{Top-3}: [Title: Mister Fantastic] Mister Fantastic (Reed Richards) is a fictional superhero appearing in American comic books published by Marvel Comics. The character is a founding member of the Fantastic Four. Richards\ldots\\
\bottomrule
\end{tabular}
\caption{Examples of DPR retrieved results on simple entity-centric questions.}
\label{tab:dpr-examples}
\end{table*}

\section{Full Results on {\eqa}}
\label{app:full_results}

\paragraph{DPR vs. BM25} The evaluation results are shown in Table~\ref{tab:full_results}.
BM25 significantly outperforms DPR models trained on either a single dataset NQ or a combination of common QA datasets.

\paragraph{REALM vs. BM25} We also evaluate he retrieval performance of REALM~\cite{guu2020realm} on {\eqa}. Specifically, we use REALM to retrieve 20 passages and check if the gold answer is a sub-string of the retrieved passages. We also evaluate BM25 on the same 288-token blocks that are used in REALM model. As shown in Table~\ref{tab:realm_results}, the results show that REALM still significantly underperforms BM25 on {\eqa}, even with the extra pre-training tasks.

\paragraph{Examples of DPR retrived passages} Table~\ref{tab:dpr-examples} shows examples of DPR retrieved results on three representative questions.
DPR makes clear mistakes like confusing entities with similar names or missing the presence of an entity, causing it to retrieve irrelevant passages on these simple, entity-centric questions.

\section{Experimental Details}
\label{app:exp-details}
\paragraph{Experimental settings of DPR}
In our experiments, we use either pre-trained DPR models released by the authors, or the DPR models re-trained by ourself (Table~\ref{tab:passage_encoders}).
All our experiments are carried out on $4 \times$ 11Gb Nvidia RTX 2080Ti GPUs.
For all our fine-tuning experiments, we fine-tune for 10 epochs, with a learning rate $2 \times 10^{-5}$ and a batch size of 24.
When we retrain DPR from scratch, we train for 20 epochs with a batch size of 24 (the original DPR models were trained on 8$\times$ 32Gb GPUs with a batch size of 128 and we have to reduce the batch size due to the limited computational resources) and a learning rate of $2 \times 10^{-5}$.

\paragraph{Experimental settings of BM25}
In our experiments, we use the Pyserini~\cite{lin2021pyserini} implementation of unigram BM25 with default parameters. We build an index using the same Wikipedia passage splits provided in the official DPR release.

\paragraph{PAQ dataset sampling}
\citet{lewis2021paq} introduce Probably Asked Questions (PAQ), a large question repository constructed using a question generation model on Wikipedia passages.
We group all of the questions asked about a particular passage and filter out any passages that have less than 3 generated questions. We then sample 100K such passages and sample one question asked about each. We split this dataset into 70K/15K/15K for train/dev/test splits, although we do not evaluate on this dataset. Following \citet{karpukhin2020dense}, we use BM25 to mine hard negative examples.

\section{Per-relation Accuracy with Different Passage Encoders}
\label{app:per-relation}
We fine-tune DPR with the passage encoder fixed on either NQ or PAQ.
Table~\ref{tab:full__passage_encoders} compares the per-relation accuracy of DPR with fixed passage encoder fine-tuned on NQ and PAQ. As is shown, the passage encoder trained on PAQ is much more robust than the passage encoder trained on NQ.
For many non-person relations, using a PAQ-based passage encoder can outperform BM25.


\begin{table*}[!t]
    \centering
    \resizebox{2\columnwidth}{!}{
    \begin{tabular}{ll | rrr | rrr | r}
        \toprule
        \multicolumn{2}{l|}{\tf{Relation}} &
        \begin{tabular}{@{}c@{}}DPR\\ (NQ)\end{tabular}
        & \begin{tabular}{@{}c@{}} Per-rel FT\\(OnlyQ)\end{tabular}
        & \begin{tabular}{@{}c@{}} EQ FT \\(OnlyQ) \end{tabular}
        & \begin{tabular}{@{}c@{}}DPR\\ (PAQ)\end{tabular}
        &  \begin{tabular}{@{}c@{}} Per-rel FT\\(OnlyQ)\end{tabular}
        & \begin{tabular}{@{}c@{}} EQ FT \\(OnlyQ) \end{tabular}
        & BM25 \\
        \midrule
P106 & What kind of work does {\entity} do? & 19.9 & 59.6 & 19.9 & 47.7 & \tf{71.6} & 47.7 & 71.2 \\
P112 & Who founded {\entity}? & 74.7 & 72.2 & 73.3 & 75.1 & 74.9 & \tf{76.3} & \tf{81.2} \\
P127 & Who owns {\entity}? & 46.5 & 70.3 & 46.5 & 63.4 & \tf{73.6} & 63.4 & \tf{78.4} \\
P131 & Where is {\entity} located? & 44.1 & 50.6 & 49.9 & 42.1 & 49.5 & \tf{50.8} & \tf{63.1} \\
P136 & What type of music does {\entity} play? & 34.7 & 57.3 & 54.8 & 44.7 & \tf{57.6} & 56.0 & 48.7 \\
P159 & Where is the headquarter of {\entity}? & 69.0 & 77.7 & 78.3 & 72.2 & 75.5 & \tf{78.4} & \tf{85.0} \\
P17 & Which country is {\entity} located in? & 56.6 & 63.9 & 64.2 & 58 & \tf{65.2} & 64.3 & 61.5 \\
P170 & Who was {\entity} created by? & 33.4 & 64.8 & 33.4 & 66.1 & \tf{75.6} & 66.1 & 72.6 \\
P175 & Who performed {\entity}? & 41.6 & 56.2 & 41.6 & 51.4 & \tf{57.8} & 51.4 & 56.6 \\
P176 & Which company is {\entity} produced by? & 43.0 & 81.0 & 43.0 & 73.9 & \tf{82.2} & 73.9 & 81.0 \\
P19 & Where was {\entity} born? & 26.0 & 48.1 & 53.8 & 54.6 & 63.9 & \tf{64.4} & \tf{75.3} \\
P20 & Where did {\entity} die? & 32.8 & 61.1 & 65.4 & 63.1 & \tf{71.8} & 70.7 & \tf{80.4} \\
P26 & Who is {\entity} married to? & 25.1 & 32.7 & 38.5 & 60.8 & \tf{69.4} & \tf{69.4} & \tf{89.7} \\
P264 & What music label is {\entity} represented by? & 27.6 & 47.9 & 27.6 & 47.1 & \tf{53.8} & 52.2 & 45.6 \\
P276 & Where is {\entity} located? & 71.4 & \tf{80.8} & 76.8 & 73.8 & 80.6 & 80.2 & \tf{84.9} \\
P36 & What is the capital of {\entity}? & 74.9 & 82.2 & 74.9 & 76.6 & 82.4 & \tf{85.7} & \tf{90.6} \\
P40 & Who is {\entity}'s child? & 16.5 & 46.0 & 16.5 & 49.7 & 63.0 & \tf{63.9} & \tf{85.0}\\
P407 & Which language was {\entity} written in? & 72.9 & 81.6 & 85.1 & 73.7 & 84.2 & \tf{86.5} & 86.2 \\
P413 & What position does {\entity} play? & 75.7 & 85.3 & 75.7 & 69.3 & \tf{85.7} & 69.3 & 74.3 \\
P495 & Which country was {\entity} created in? & 19.4 & 24.2 & \tf{35.1} & 20.2 & 26.0 & 30.0 & 21.8 \\
P50 & Who is the author of {\entity}? & 75.7 & 79.4 & 79.6 & 74.8 & \tf{79.8} & 79.2 & 73.0 \\
P69 & Where was {\entity} educated? & 19.9 & 55.9 & 55.8 & 48.9 & 68.1 & \tf{69.5} & \tf{73.1} \\
P740 & Where was {\entity} founded? & 57.0 & 77.8 & 57.0 & 67.6 & \tf{79.9} & 67.6 & 74.4 \\
P800 & What is {\entity} famous for? & 24.4 & 22.6 & 25.8 & 49.3 & 49.8 & \tf{52.0} & \tf{74.7} \\
        \midrule
        \multicolumn{2}{l|}{\ti{Macro-Average}} & 45.1 & 61.6 & 53.0 & 59.3 & \tf{68.4} & 65.4 & \tf{72.0} \\
        \multicolumn{2}{l|}{\ti{Micro-Average}} & 44.6 & 62.3 & 53.0 & 59.0 & \tf{68.5} & 65.1 & \tf{71.4} \\
        \bottomrule
    \end{tabular}
    }
    \caption{Top-20 retrieval accuracy on NQ and {\eqa} (EQ). \textit{Per-rel FT}: we fine-tune an individual question encoder for each relation. \textit{EQ FT}: we fine-tune a single question encoder on all relations in {\eqa}.
    }
    \label{tab:full__passage_encoders}
\end{table*}

\end{document}